\documentclass[letterpaper,10pt,conference]{cssconf}

\IEEEoverridecommandlockouts

\usepackage{graphicx}
\usepackage{amsmath}
\usepackage{amssymb}
\usepackage{cite}
\usepackage{booktabs}
\usepackage{array}
\usepackage{xcolor}
\usepackage{url}

\newcommand{\R}{\mathbb{R}}
\newcommand{\norm}[1]{\left\lVert #1\right\rVert}
\newcommand{\vect}[1]{\boldsymbol{#1}}
\newcommand{\dd}{\mathrm{d}}
\newtheorem{theorem}{Theorem}

\title{Safe Physical Interaction of All-Electric Heavy-Duty Robotic Manipulators via Control Barrier Functions and Variable Impedance Control%
\thanks{This work was supported by the Research Council of Finland under the project ``Nonlinear PDE-model-based control of flexible manipulators'' under Grant 355664. 
This work has been submitted to the IEEE for possible
publication. Copyright may be transferred without notice, after which this
version may no longer be accessible.}
}

\author{
\authorblockN{Amir Hossein Barjini, Mohammad Bahari, Jouni Mattila}
\authorblockA{
Faculty of Engineering and Natural Sciences, Tampere University, 33720 Tampere, Finland\\
Email: amirhossein.barjini@tuni.fi
}
}

\begin{document}
\maketitle
\thispagestyle{empty}
\pagestyle{empty}

\begin{abstract}
All-electric heavy-duty robotic manipulators are increasingly expected to operate close to tunnel walls, equipment, and other structures while retaining the force capability required for large-payload tasks. Accurate motion control alone does not address the resulting safety and interaction requirements. This paper develops a safety-aware physical-interaction architecture that combines variable impedance control (VIC) with control barrier functions (CBFs) on top of an established virtual decomposition control (VDC) framework for electromechanical linear actuator (EMLA)-driven manipulators. VIC continuously adapts the task-space stiffness and damping according to predicted proximity and contact risk, whereas a joint-velocity CBF filter enforces Cartesian workspace boundaries and the physical stroke limits induced by the closed-chain lift and tilt mechanisms. To account for nonideal realization of the filtered velocity by the inner VDC/EMLA dynamics, the CBF inequalities are tightened using a measured bound on the joint-velocity tracking error. Four configurations, namely VDC, VIC+VDC, CBF+VDC, and VIC+CBF+VDC, are evaluated under workspace, actuator-limit, and physical-interaction scenarios. VIC reduces the peak and RMS environment-interaction forces by \(92.59\%\) and \(96.33\%\), respectively; both robust CBF configurations maintain positive workspace and EMLA safety margins, while VIC+CBF+VDC reduces the CBF intervention metric by \(10.30\%\) compared with CBF+VDC.
\end{abstract}


\section{Introduction}
Heavy-duty robotic manipulators (HDRMs) are increasingly used in drilling, mining, construction, forestry, and material handling, where long reach and large payload capacity are required under demanding operating conditions. Achieving reliable and adaptable manipulation in such environments remains challenging because of nonlinear dynamics, strong actuator--structure coupling, and large interaction forces \cite{billard2019trends,li2019compliance}. Electrification is increasingly reshaping the actuation architectures of heavy-duty mobile machines, with recent studies investigating electric and electro-hydraulic actuation concepts to improve energy efficiency, enable energy recovery, and enhance actuator-level controllability \cite{fassbender2021minav,fresia2022combined,qu2023electrified}. In fully electric implementations, an electromechanical linear actuator (EMLA) combines an electric motor with a mechanical transmission, typically a gearbox and screw mechanism, to generate the linear force required by manipulator joints. Such actuator-level electrification can eliminate hydraulic throttling and enable more direct integration of actuation, sensing, and control, but it also introduces actuator-dependent force, velocity, and stroke constraints that must be considered at the manipulator level \cite{bahari2026all,barjini2025surrogate,bahari2026structure}. Consequently, safe operation of all-electric HDRMs requires not only accurate motion realization but also explicit treatment of actuator and workspace constraints.

Control of HDRMs must address nonlinear multibody dynamics, varying loads, actuator constraints, and model uncertainty \cite{hejrati2026adaptive}. For hydraulically actuated systems, nonlinear model predictive control (NMPC), data-driven control, and reduced-order task-space NMPC have been investigated for trajectory tracking, obstacle avoidance, and actuator-aware motion \cite{mononen2019nmpc,yao2023data,kowsari2026reduced}. For all-electric HDRMs, virtual decomposition control (VDC) provides a modular model-based framework that has recently been extended to coordinate manipulator and electromechanical actuator dynamics \cite{zhu2010vdc,bahari2026all,barjini2025surrogate}. These approaches improve tracking and command realization, but safety near workspace and actuator limits still requires explicit constraint enforcement.

For admissible physical interaction, the controller must additionally regulate the robot's mechanical response to contact. Impedance control shapes the dynamic relationship between robot motion and interaction forces \cite{hogan1985impedance}, while variable impedance control (VIC) adapts the desired impedance, particularly stiffness and damping, as the task or environment changes \cite{abudakka2020vicreview}. Since time-varying impedance parameters can affect closed-loop stability, their adaptation requires careful treatment \cite{kronander2016stability}. Desired impedance allocation has also been incorporated into the modular VDC framework for interaction control \cite{hejrati2025impedanceallocation}. In parallel, sensor-based perception has been shown to support context-aware human--robot collaboration in industrial tasks \cite{angleraud2024sensor}. Nevertheless, impedance modulation alone does not provide explicit guarantees on hard geometric or actuator constraints.

Control barrier functions (CBFs) provide a systematic means of enforcing safety by rendering a prescribed safe set forward invariant under appropriate conditions. When incorporated as constraints in quadratic programs (QPs), CBFs can reconcile nominal control objectives with explicit safety requirements \cite{ames2017cbfqp}. Their use has been extended to robotic kinematics, including velocity-based safety-aware control of manipulators \cite{singletary2022kinematic}. CBF-QP safety filtering has also been applied to collision-aware motion of long-reach drilling booms \cite{shahna2026pose}, and to actuator-aware inverse kinematics of redundant robots \cite{dastranj2026actuator}. Robust CBF formulations further address disturbances and perturbations that can affect preservation of the safe set \cite{xu2015robustcbf}. Barrier-based safety has also been integrated with impedance control for constrained robotic interaction \cite{wang2022high}, while predictive impedance adaptation combined with safety CBFs has been investigated for obstacle avoidance in human--robot collaboration \cite{ducaju2025predictive}. These studies demonstrate the compatibility of impedance-based interaction control and barrier-based safety, motivating an actuator-aware formulation for all-electric HDRMs.

\begin{figure}[ht]
\centering
\includegraphics[width=0.94\columnwidth]{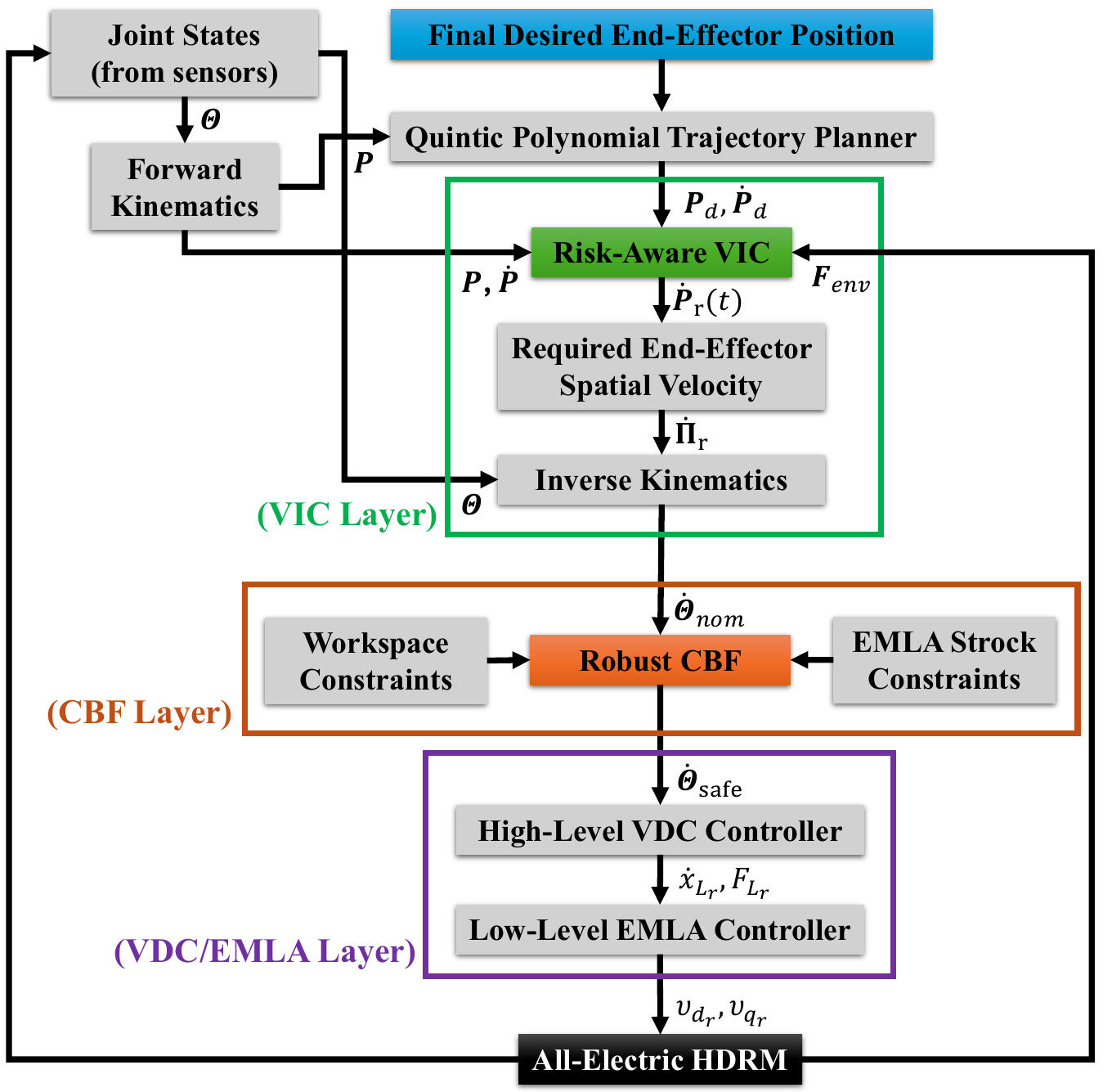}
\caption{Proposed VIC--CBF--VDC control architecture for safe physical interaction of the all-electric HDRM.}
\label{fig:VIC_CBF_VDC_Schematic}
\end{figure}

Despite recent progress in all-electric HDRM modeling and control, safe physical interaction remains insufficiently addressed when workspace constraints and closed-chain actuator limits must be enforced simultaneously. Building on the analytical EMLA/HDRM models and modular control framework in \cite{bahari2026all,barjini2025surrogate}, this paper develops an actuator-aware VIC--CBF--VDC architecture. The main contributions are:
\begin{enumerate}
\item a hierarchical framework in which VIC shapes interaction behavior, the CBF layer enforces safety, and VDC with low-level EMLA control executes the resulting commands;
\item a unified CBF formulation for Cartesian workspace constraints and closed-chain lift/tilt EMLA stroke limits;
\item robust CBF tightening based on a measured bound on inner-loop joint-velocity tracking error; and
\item comparison of VDC, VIC+VDC, CBF+VDC, and VIC+CBF+VDC using tracking, safety, intervention, and interaction metrics.
\end{enumerate}
The overall control architecture is illustrated in Fig.~\ref{fig:VIC_CBF_VDC_Schematic}, showing the information flow from the desired end-effector motion and measured joint states through the risk-aware VIC, robust CBF safety filter, high-level VDC, and low-level EMLA controller to the all-electric HDRM.

The remainder of the paper is organized as follows. Section II presents the HDRM, VDC, and EMLA models. Section III introduces the VIC and CBF framework, and Section IV provides the stability and safety analysis. Section V presents the validation setup and results, and Section VI concludes the paper.

\section{System Modeling and Baseline Control}

\subsection{All-Electric Heavy-Duty Robotic Manipulator}
\begin{figure}[t]
\centering
\includegraphics[width=0.9\columnwidth]{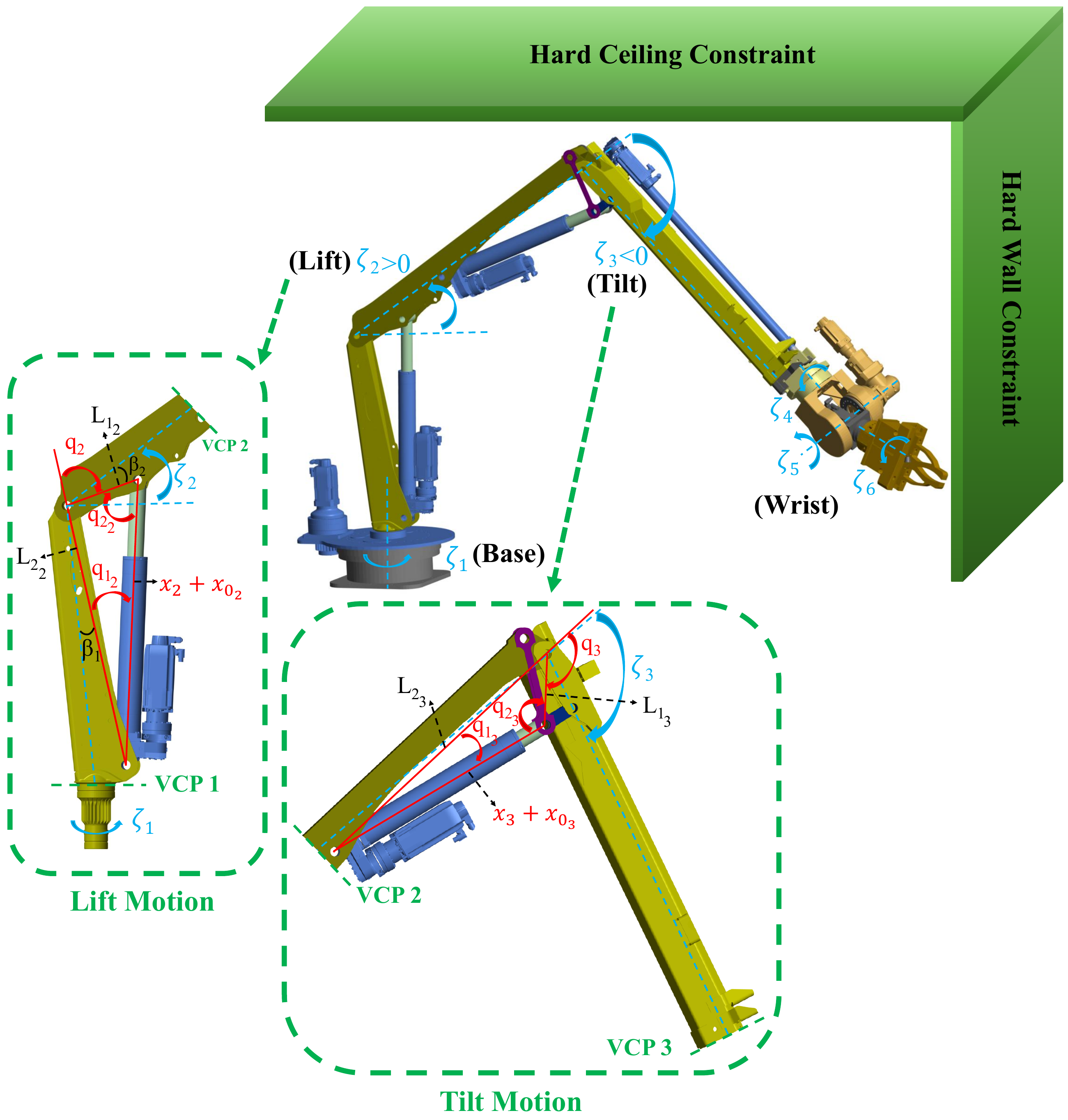}
\caption{All-electric HDRM and safety scenario with hard wall/ceiling constraints.}
\label{fig:manipulator_structure}
\end{figure}

The considered system is the six-DoF all-electric HDRM studied in \cite{bahari2026all,barjini2025surrogate}. As illustrated in Fig.~\ref{fig:manipulator_structure}, its serial-chain configuration consists of the base, lift, tilt, and three-DoF wrist, while the lift and tilt motions are realized through closed-chain mechanisms driven by EMLAs. The figure also illustrates the corresponding lift and tilt geometries together with the hard ceiling and wall constraints considered in the safety formulation. Define
\begin{equation}
\vect{\Theta}=\begin{bmatrix}\zeta_1&\zeta_2&\zeta_3&\zeta_4&\zeta_5&\zeta_6\end{bmatrix}^{\!T}\in\R^6
\label{eq:Theta}
\end{equation}
as the joint-angle vector and \(\vect P=[x\;y\;z]^T\) as the end-effector position. Let \(\dot{\vect\Pi}=[\dot{\vect P}^{T}\;\vect\omega_e^{T}]^{T}\in\R^6\) denote the end-effector spatial velocity. The manipulator kinematics are written as
\begin{equation}
\dot{\vect\Pi}=\vect J(\vect{\Theta})\dot{\vect{\Theta}},\qquad
\dot{\vect P}=\vect J_P(\vect{\Theta})\dot{\vect{\Theta}},
\label{eq:jac}
\end{equation}
where \(\vect J\in\R^{6\times6}\) is the full Jacobian and \(\vect J_P\in\R^{3\times6}\) is its translational part. The safety layer developed in this paper acts on the end-effector position and on the lift/tilt EMLA strokes, while the complete six-DoF kinematic and dynamic model is retained in the inner controller.

The lift and tilt EMLA extensions are denoted by \(x_2\) and \(x_3\), respectively \cite{bahari2026all,barjini2025surrogate}. For compact barrier notation, define \(s_L:=x_2\) and \(s_T:=x_3\). For the lift three-bar mechanism, the geometric angle satisfies
\begin{equation}
q_2=\left(-\frac{\pi}{2}-\beta_1-\beta_2\right)+\zeta_2,
\label{eq:liftangle}
\end{equation}
and the EMLA extension follows from the law of cosines,
\begin{equation}
 s_L=\sqrt{L_{1_2}^2+L_{2_2}^2+2L_{1_2}L_{2_2}\cos q_2}-x_{0_2}.
\label{eq:liftstroke}
\end{equation}
Since \(\dot q_2=\dot\zeta_2\), differentiation gives
\begin{equation}
 \dot s_L=g_L(\zeta_2)\dot\zeta_2,\qquad
 g_L=-\frac{L_{1_2}L_{2_2}\sin q_2}{s_L+x_{0_2}}.
\label{eq:liftg}
\end{equation}
For the tilt four-bar mechanism, the established geometric relations in \cite{bahari2026all,barjini2025surrogate} map \(\zeta_3\) to \(L_{1_3}(\zeta_3)\) and \(q_3(\zeta_3)\). The corresponding extension is
\begin{equation}
 s_T=\sqrt{L_{1_3}^2+L_{2_3}^2+2L_{1_3}L_{2_3}\cos q_3}-x_{0_3},
\label{eq:tiltstroke}
\end{equation}
and its exact transmission ratio is written compactly as
\begin{equation}
 \dot s_T=g_T(\zeta_3)\dot\zeta_3,\qquad
 g_T(\zeta_3):=\frac{\partial x_3(\zeta_3)}{\partial\zeta_3}.
\label{eq:tiltg}
\end{equation}
Thus, the stroke constraints later act through the actual closed-chain transmission ratios rather than through approximate rotary-joint limits.

The baseline manipulator controller follows a hierarchical structure composed of a high-level VDC layer and low-level EMLA controllers. The high-level controller computes the required motion and actuator force commands for the manipulator subsystems, while the low-level electromechanical controllers realize these requirements through the PMSM-driven EMLAs.

\subsection{High-Level Virtual Decomposition Control}
The high-level controller follows the VDC formulation used in \cite{zhu2010vdc,bahari2026all,barjini2025surrogate}. VDC decomposes the manipulator into rigid-body and actuator subsystems connected through virtual cutting points (VCPs). For a body-fixed frame \(\{A\}\), the spatial velocity and spatial force are defined as
\begin{equation}
{}^{A}\vect V=\begin{bmatrix}{}^{A}\vect v\\{}^{A}\vect\omega\end{bmatrix},\qquad
{}^{A}\vect F=\begin{bmatrix}{}^{A}\vect f\\{}^{A}\vect\tau\end{bmatrix}\in\R^6,
\label{eq:vdcVF}
\end{equation}
where \({}^{A}\vect v\) and \({}^{A}\vect\omega\) denote the linear and angular velocity components, respectively, while \({}^{A}\vect f\) and \({}^{A}\vect\tau\) denote the force and moment components of the spatial force. For another frame \(\{B\}\) attached to the same rigid body, the VDC transformations are
\begin{equation}
{}^{B}\vect V={}^{A}\vect U_B^{T}{}^{A}\vect V,\qquad
{}^{A}\vect F={}^{A}\vect U_B{}^{B}\vect F,
\label{eq:vdcTransform}
\end{equation}
where \({}^{A}\vect U_B\in\R^{6\times6}\) is the spatial transformation matrix defined by the relative rotation and displacement of the two frames \cite{zhu2010vdc}. Accordingly, spatial velocities are propagated recursively from the base toward the end-effector, whereas spatial forces are propagated from the end-effector toward the base \cite{bahari2026all,barjini2025surrogate}.

For each rigid body, the net spatial force can be expressed in the regressor form
\begin{equation}
\vect M_A\,{}^{A}\dot{\vect V}+\vect C_A{}^{A}\vect V+\vect G_A
={}^{A}\vect F^{*}={}^{A}\vect Y\vect\theta_A,
\label{eq:vdcdyn}
\end{equation}
where \(\vect M_A\), \(\vect C_A\), and \(\vect G_A\) are the spatial inertia, Coriolis/centrifugal, and gravity terms, respectively. The required spatial force is selected as
\begin{equation}
{}^{A}\vect F_r^{*}={}^{A}\vect Y\hat{\vect\theta}_{A}
+{}^{A}\vect K_A\left({}^{A}\vect V_r-{}^{A}\vect V\right),
\label{eq:vdcwrench}
\end{equation}
with \({}^{A}\vect K_A \succ 0\). The required spatial velocity \({}^{A}\vect V_r\) is computed through the same forward-recursion structure used for the actual spatial velocity, with the measured joint and closed-chain rates replaced by their corresponding required values. Based on these required velocities, the required spatial force is evaluated for each subsystem and subsequently propagated backward through the VCPs. This backward recursion ultimately yields the required rotary-joint torques and EMLA forces. The natural adaptation laws adopted in \cite{bahari2026all,barjini2025surrogate} update the uncertain inertial parameters online while retaining the modular structure of the VDC framework.

\subsection{Low-Level Electromechanical Linear Actuator Control}
At the low level of the hierarchical controller, each EMLA consists of a PMSM, a reduction gearbox, and a screw mechanism. For a balanced three-phase PMSM, the current dynamics and electromagnetic torque in the rotating \(dq\) frame are expressed as \cite{bahari2026all,barjini2025surrogate}
\begin{equation}
\begin{aligned}
\dot i_d&=\frac{v_d-r_si_d+L_qP\omega_m i_q}{L_d},\\
\dot i_q&=\frac{v_q-r_si_q-L_dP\omega_m i_d-P\psi_f\omega_m}{L_q},\\
\tau_e&=\frac{3}{2}P\left[\psi_f i_q+(L_d-L_q)i_di_q\right],
\end{aligned}
\label{eq:pmsm}
\end{equation}
where \(P\) is the pole-pair number, \(\omega_m\) is the motor angular velocity, \(\psi_f\) is the permanent-magnet flux linkage, and \(L_d\), \(L_q\), and \(r_s\) denote the \(dq\)-axis inductances and stator resistance, respectively.

The motor torque is transmitted to the load through the gearbox and screw mechanism. With \(n=2\pi/\rho\), where \(\rho\) is the screw lead, define the direction-dependent transmission factors as
\begin{equation}
\kappa_f=nN_g\eta_t^{+},
\qquad
\kappa_b=\frac{nN_g\eta_g}{\eta_b},
\label{eq:kappa}
\end{equation}
where \(N_g\) is the gearbox ratio, \(\eta_t^{+}\) is the forward transmission efficiency, and \(\eta_g\) and \(\eta_b\) denote the gearbox and backward screw efficiencies, respectively. For compactness, let
\begin{equation}
\mu_f=M_t+nN_g\kappa_fJ_m,
\qquad
\mu_b=M_t+nN_g\kappa_bJ_m.
\label{eq:effective_mass}
\end{equation}
The EMLA load-side force can then be obtained from the analytical actuator dynamics as \cite{bahari2026all}
\begin{equation}
F_L=
\begin{cases}
\kappa_f(\tau_e-\tau_c)-f_v\dot x_L
-\mu_f\ddot x_L,
& \dot x_L\geq0,\\[3pt]
\kappa_b(\tau_e-\tau_c)-f_v\dot x_L
-\mu_b\ddot x_L,
& \dot x_L<0,
\end{cases}
\label{eq:loadforce}
\end{equation}
where \(\tau_c\), \(f_v\), \(J_m\), and \(M_t\) denote the Coulomb friction, viscous damping coefficient, motor inertia, and total moving mass, respectively. Here, \(F_L\) represents the EMLA load-side force and is distinct from the Cartesian environment interaction force \(\vect F_{\mathrm{env}}\) used in the variable impedance control layer.

The high-level VDC controller provides the required load-side force \(F_{L_r}\) and linear velocity \(\dot x_{L_r}\) for each EMLA. These references are realized by the low-level controller through the required \(dq\)-axis voltages \cite{bahari2026all,barjini2025surrogate},
\begin{equation}
\begin{aligned}
v_{d_r}={}&r_si_d+L_d\frac{di_{d_r}}{dt}
-P\omega_m\lambda_q
+K_i(i_{d_r}-i_d),\\
v_{q_r}={}&r_si_q+L_q\frac{di_{q_r}}{dt}
+P\omega_m\lambda_d\\
&+K_f(F_{L_r}-F_L)
+K_v(\dot x_{L_r}-\dot x_L),
\end{aligned}
\label{eq:lowlevel}
\end{equation}
where
\(\lambda_d=L_di_d+\psi_f\) and
\(\lambda_q=L_qi_q\) are the \(dq\)-axis flux linkages, while \(K_i\), \(K_f\), and \(K_v\) are positive current, force, and velocity feedback gains, respectively. The required \(d\)-axis current is generated according to
\begin{equation}
\frac{di_{d_r}}{dt}
=
\frac{di_{d_d}}{dt}
+\lambda_i(i_{d_d}-i_d),
\label{eq:idref}
\end{equation}
where \(\lambda_i>0\), and \(i_{d_d}=0\) can be selected for the conventional efficient PMSM operating condition. The required \(q\)-axis current is obtained from the required actuator force through the motor--transmission relationship~\cite{barjini2025surrogate,bahari2026all}.

Consequently, the high-level VDC computes the required EMLA force and velocity, while the low-level controller converts these requirements into the PMSM voltage commands \(v_{d_r}\) and \(v_{q_r}\). Together, these two layers constitute the hierarchical baseline controller used beneath the VIC and CBF safety layers.

\section{safety-aware Variable Impedance Control}

\subsection{Risk-Aware Variable Impedance Reference}
Let \(\vect P_d,\dot{\vect P}_d\) denote the desired end-effector position and velocity, and let \(\vect F_d\) and \(\vect F_{\mathrm{env}}\) denote the desired and measured environment interaction forces, respectively. The variable impedance relation is defined as
\begin{equation}
\vect D_d(\dot{\vect P}_r-\dot{\vect P}_d)
+\vect K_d(\vect P-\vect P_d)
=\vect F_d-\vect F_{\mathrm{env}},
\label{eq:vic}
\end{equation}
which gives the required end-effector velocity
\begin{equation}
\dot{\vect P}_r=
\dot{\vect P}_d+
\vect D_d^{-1}\vect K_d(\vect P_d-\vect P)
+\vect D_d^{-1}(\vect F_d-\vect F_{\mathrm{env}}).
\label{eq:vicref}
\end{equation}
The diagonal positive-definite stiffness and damping matrices are scheduled by a scalar risk index \(\chi\in[0,1]\):
\begin{align}
\vect K_d(\chi)&=(1-\chi)\vect K_{\mathrm{far}}
+\chi\vect K_{\mathrm{near}},\\
\vect D_d(\chi)&=(1-\chi)\vect D_{\mathrm{far}}
+\chi\vect D_{\mathrm{near}}.
\label{eq:impinterp}
\end{align}

For the wall and ceiling, the physical clearances are defined as
\begin{equation}
d_x=x_{\mathrm{wall}}-x,\qquad
d_y=y_{\mathrm{ceiling}}-y.
\end{equation}
Only motion toward a boundary contributes to the predicted risk. Accordingly,
\begin{equation}
v_{c,x}=\max(\dot x,0),\qquad
v_{c,y}=\max(\dot y,0),
\end{equation}
and the predicted clearance is
\begin{equation}
\hat d_j=d_j-T_pv_{c,j},
\qquad j\in\{x,y\},
\label{eq:predict}
\end{equation}
where \(T_p\) denotes the prediction horizon.

The proximity risks associated with the wall and ceiling are defined as
\begin{align}
\chi_x &=
S\!\left(
\operatorname{sat}_{[0,1]}
\frac{d_{\mathrm{warn}}-\hat d_x}
{d_{\mathrm{warn}}-d_{\mathrm{env}}}
\right),\\
\chi_y &=
S\!\left(
\operatorname{sat}_{[0,1]}
\frac{d_{\mathrm{warn}}-\hat d_y}
{d_{\mathrm{warn}}-d_{\mathrm{env}}}
\right),
\label{eq:proxrisk}
\end{align}
where \(\operatorname{sat}_{[0,1]}(\cdot)\) denotes saturation to the interval \([0,1]\). The parameter \(d_{\mathrm{warn}}\) is the warning distance at which VIC begins modifying the nominal impedance, whereas \(d_{\mathrm{env}}\) is the smaller environmental safety margin defining the CBF boundary, with
\(d_{\mathrm{warn}}>d_{\mathrm{env}}>0\).
The mapping \(S(\cdot)\) is selected as the cubic smoothstep function
\begin{equation}
S(r)=3r^2-2r^3,
\label{eq:smoothstep}
\end{equation}
which satisfies \(S(0)=0\), \(S(1)=1\), and \(S'(0)=S'(1)=0\). Hence, the impedance parameters transition smoothly between the far- and near-boundary settings without abrupt changes at the endpoints of the scheduling interval.

Similarly, the interaction-force risk is defined as
\begin{equation}
\chi_F=
S\!\left(
\operatorname{sat}_{[0,1]}
\frac{\|\vect F_{\mathrm{env}}\|-F_{\mathrm{on}}}
{F_{\mathrm{full}}-F_{\mathrm{on}}}
\right),
\label{eq:forcerisk}
\end{equation}
where \(F_{\mathrm{on}}\) denotes the force magnitude at which the environment interaction begins influencing the impedance, while \(F_{\mathrm{full}}>F_{\mathrm{on}}\) corresponds to full impedance adaptation. The overall risk index is then selected as
\begin{equation}
\chi=\max\{\chi_x,\chi_y,\chi_F\}.
\label{eq:riskmax}
\end{equation}
Thus, VIC gradually adapts the nominal interaction behavior as the end-effector approaches an environmental boundary or as the environment interaction force increases. VIC itself, however, does not impose a hard safety constraint.

For comparison, when VIC is disabled, the original VDC position reference used in \cite{bahari2026all} is retained,
\begin{equation}
\dot{\vect P}_r^{0}
=
\dot{\vect P}_d+
\lambda_p(\vect P_d-\vect P),
\qquad \lambda_p>0,
\label{eq:originalref}
\end{equation}
where \(\lambda_p\) is the position-feedback gain. As in the baseline controller, the present task regulates the end-effector position while maintaining a fixed orientation; hence, the required angular-velocity vector is set to zero. The required spatial velocity is therefore written as
\begin{equation}
\dot{\vect\Pi}_r=
\begin{bmatrix}
\dot{\vect P}_r^T & \vect\omega_{e,r}^T
\end{bmatrix}^{T},
\qquad
\vect\omega_{e,r}=\vect 0,
\label{eq:requiredspatialvelocity}
\end{equation}
where \(\dot{\vect P}_r\) is given by \eqref{eq:originalref} when VIC is disabled and by \eqref{eq:vicref} when VIC is enabled. Both references are mapped through the same inverse kinematics and tracked by the same hierarchical VDC/EMLA controller. When enabled, the CBF filter described next modifies the resulting joint-velocity command only as required by the workspace or EMLA stroke constraints. Thus, VIC shapes the nominal interaction behavior, whereas the CBF layer remains responsible for hard constraint enforcement. Because the CBF acts in joint-velocity space, Cartesian workspace and closed-chain EMLA stroke constraints can be handled simultaneously; feasibility requires the intersection of the resulting inequalities to remain nonempty and is monitored by the active-set solver.

\subsection{Control Barrier Functions}

\subsubsection{Workspace Control Barrier Functions}
The CBF boundary is placed inside the physical wall and ceiling by the environmental safety margin \(d_{\mathrm{env}}\):
\begin{equation}
x_{\max}^{s}=x_{\mathrm{wall}}-d_{\mathrm{env}},
\qquad
y_{\max}^{s}=y_{\mathrm{ceiling}}-d_{\mathrm{env}}.
\label{eq:safeworkspace}
\end{equation}
The corresponding barrier functions are
\begin{equation}
h_x=x_{\max}^{s}-x,\qquad
h_y=y_{\max}^{s}-y.
\label{eq:envh}
\end{equation}
Thus, \(h_x,h_y>0\) denotes the interior of the safe workspace, \(h_i=0\) the designed safety boundary, and \(h_i<0\) a constraint violation.

Using the zeroing-CBF condition
\(\dot h_i+\gamma_i h_i\geq0\), with \(\gamma_i>0\), and the kinematics in \eqref{eq:jac}, the workspace constraints become
\begin{equation}
\vect J_x\dot{\vect\Theta}\leq\gamma_xh_x,\qquad
\vect J_y\dot{\vect\Theta}\leq\gamma_yh_y,
\label{eq:envcbf}
\end{equation}
where \(\vect J_x\) and \(\vect J_y\) are the first and second rows of \(\vect J_P\).

\subsubsection{EMLA Stroke Barrier Functions}
Let the physical lift and tilt stroke intervals be
\(
[s_{L,\min},s_{L,\max}],
\qquad
[s_{T,\min},s_{T,\max}],
\)
respectively, and let \(d_{\mathrm{stroke}}>0\) denote the mechanical safety margin. The corresponding safe stroke limits are
\begin{align}
s_{L,\min}^{s}&=s_{L,\min}+d_{\mathrm{stroke}},&
s_{L,\max}^{s}&=s_{L,\max}-d_{\mathrm{stroke}},\\
s_{T,\min}^{s}&=s_{T,\min}+d_{\mathrm{stroke}},&
s_{T,\max}^{s}&=s_{T,\max}-d_{\mathrm{stroke}}.
\label{eq:safestroke}
\end{align}
The four stroke barrier functions are then defined as
\begin{align}
h_{L,\min}&=s_L-s_{L,\min}^{s},&
h_{L,\max}&=s_{L,\max}^{s}-s_L,\\
h_{T,\min}&=s_T-s_{T,\min}^{s},&
h_{T,\max}&=s_{T,\max}^{s}-s_T.
\label{eq:strokeh}
\end{align}
Each barrier therefore represents the remaining distance to one safe stroke boundary. Positive values indicate operation inside the admissible interval.

Using the closed-chain transmission relations \eqref{eq:liftg} and \eqref{eq:tiltg}, the CBF conditions yield
\begin{align}
-g_L\dot\zeta_2&\leq\gamma_{L,\min}h_{L,\min},&
g_L\dot\zeta_2&\leq\gamma_{L,\max}h_{L,\max},\\
-g_T\dot\zeta_3&\leq\gamma_{T,\min}h_{T,\min},&
g_T\dot\zeta_3&\leq\gamma_{T,\max}h_{T,\max}.
\label{eq:strokecbf}
\end{align}
Unlike approximate joint-angle constraints, these inequalities explicitly account for the instantaneous lift and tilt transmission ratios.

The workspace and stroke constraints are collected as
\begin{equation}
\vect A(\vect\Theta)\dot{\vect\Theta}
\leq
\vect b(\vect\Theta),
\label{eq:stacked}
\end{equation}
with
\begin{equation}
\vect A=
\begin{bmatrix}
\vect J_x\\
\vect J_y\\
-g_L\vect e_2^T\\
 g_L\vect e_2^T\\
-g_T\vect e_3^T\\
 g_T\vect e_3^T
\end{bmatrix},
\quad
\vect b=
\begin{bmatrix}
\gamma_xh_x\\
\gamma_yh_y\\
\gamma_{L,\min}h_{L,\min}\\
\gamma_{L,\max}h_{L,\max}\\
\gamma_{T,\min}h_{T,\min}\\
\gamma_{T,\max}h_{T,\max}
\end{bmatrix},
\label{eq:Ab}
\end{equation}
where \(\vect e_j\in\R^6\) is the \(j\)th canonical unit vector.

\subsubsection{Robust CBF Velocity Filter}
For the nonsingular configurations considered in this work, the required end-effector velocity is mapped to the nominal joint velocity through inverse kinematics,
\begin{equation}
\dot{\vect\Theta}_{\mathrm{nom}}
=
\vect J^{-1}(\vect\Theta)\dot{\vect\Pi}_r.
\label{eq:nominalik}
\end{equation}
The CBF filter then computes the closest admissible joint-velocity command satisfying the workspace and EMLA stroke constraints:
\begin{equation}
\begin{aligned}
\dot{\vect\Theta}_{\mathrm{safe}}
=&\arg\min_{\vect u\in\R^6}
\frac{1}{2}
\|\vect u-\dot{\vect\Theta}_{\mathrm{nom}}\|_{\vect W}^{2}\\
&\mathrm{s.t.}\quad
\vect A_i\vect u\leq b_i^{r},
\qquad i=1,\ldots,6,
\end{aligned}
\label{eq:qp}
\end{equation}
where \(\vect W\succ0\) is a weighting matrix, selected as the identity matrix in the present implementation.

Because the hierarchical VDC/EMLA controller realizes the filtered command with finite velocity-tracking error, define
\begin{equation}
\vect e_v=
\dot{\vect\Theta}_{\mathrm{actual}}
-\dot{\vect\Theta}_{\mathrm{safe}},
\qquad
\|\vect e_v\|_2\leq\bar e_v,
\label{eq:ev}
\end{equation}
where \(\bar e_v\) denotes an upper bound on the joint-velocity tracking error. To account for this error, each CBF constraint is tightened as
\begin{equation}
b_i^{r}
=
b_i-\|\vect A_i\|_2\bar e_v.
\label{eq:tight}
\end{equation}
The bound \(\bar e_v\) is obtained from the closed-loop velocity-tracking response and fixed before the robust verification run. Consequently, the safety filter accounts for the finite tracking accuracy of the inner hierarchical controller rather than assuming ideal realization of \(\dot{\vect\Theta}_{\mathrm{safe}}\). The resulting safety condition is discussed together with the closed-loop stability in Section~\ref{sec:stability}.

\section{Stability and Safety Analysis}
\label{sec:stability}

The stability of the hierarchical VDC/EMLA controller is established in \cite{barjini2025surrogate} using the virtual-stability framework of VDC and the corresponding actuator-level Lyapunov analysis. The VIC and CBF layers introduced here do not modify the VDC force recursion or the low-level EMLA controller; they only modify the required velocity supplied to this established inner-loop architecture. The following analysis therefore addresses the added VIC dynamics and the robust CBF safety property.

Define the end-effector position error and velocity realization error as
\begin{equation}
\vect e_P=\vect P_d-\vect P,
\qquad
\tilde{\vect v}_P=\dot{\vect P}_r-\dot{\vect P}.
\label{eq:vicerrors}
\end{equation}
Using \eqref{eq:vicref}, the position-error dynamics are
\begin{equation}
\dot{\vect e}_P
=-\vect G_P(\chi)\vect e_P+\vect\xi_P,
\label{eq:vicerr}
\end{equation}
where
\begin{equation}
\begin{aligned}
\vect G_P(\chi)&=\vect D_d^{-1}(\chi)\vect K_d(\chi),\\
\vect\xi_P&=\vect D_d^{-1}(\chi)
(\vect F_{\mathrm{env}}-\vect F_d)
+\tilde{\vect v}_P .
\end{aligned}
\label{eq:vicdist}
\end{equation}
Since \(\vect K_d(\chi)\) and \(\vect D_d(\chi)\) are diagonal and positive definite for \(\chi\in[0,1]\), there exists \(\underline g>0\) such that \(\lambda_{\min}(\vect G_P)\geq\underline g\).

\begin{theorem}[VIC stability and robust safety]
\label{thm:vic_cbf}
Assume that the scheduled stiffness and damping matrices remain positive definite, the CBF constraints are feasible, the kinematic maps are locally Lipschitz, the joint-velocity realization error satisfies \(\|\vect e_v\|_2\leq\bar e_v\), and the initial state satisfies all six barrier conditions \(h_i(0)\geq0\). Then the VIC position-error dynamics are input-to-state stable with respect to \(\vect\xi_P\), and the robust CBF filter renders the safe set
\begin{equation}
\mathcal C=\{\vect\Theta:h_i(\vect\Theta)\geq0,\ i=1,\ldots,6\}
\label{eq:safeset}
\end{equation}
forward invariant.
\end{theorem}

\emph{Proof:}
Consider \(V_P=\tfrac12\vect e_P^T\vect e_P\). From \eqref{eq:vicerr},
\begin{align}
\dot V_P
&\leq-\underline g\|\vect e_P\|_2^2
+\|\vect e_P\|_2\|\vect\xi_P\|_2 \nonumber\\
&\leq-\frac{\underline g}{2}\|\vect e_P\|_2^2
+\frac{1}{2\underline g}\|\vect\xi_P\|_2^2,
\label{eq:viciss}
\end{align}
which establishes input-to-state stability of the VIC error dynamics. For the safety filter,
\(\dot{\vect\Theta}_{\mathrm{actual}}=\dot{\vect\Theta}_{\mathrm{safe}}+\vect e_v\). Using \eqref{eq:tight} and Cauchy--Schwarz,
\begin{equation}
\vect A_i\dot{\vect\Theta}_{\mathrm{actual}}
\leq b_i^r+\|\vect A_i\|_2\bar e_v=b_i.
\label{eq:robustproof}
\end{equation}
Thus, each actual constraint satisfies \(\dot h_i+\gamma_i h_i\geq0\), yielding
\(h_i(t)\geq h_i(0)e^{-\gamma_i t}\geq0\). Hence, \(\mathcal C\) is forward invariant. \hfill\(\square\)

In free motion away from active safety constraints, when \(\vect F_{\mathrm{env}}=\vect F_d\) and the CBF filter is inactive, the inner-loop velocity error converges to zero, \(\vect\xi_P\rightarrow0\) and consequently \(\vect e_P\rightarrow0\). During admissible physical interaction or active safety filtering, a bounded nonzero position error may remain because interaction and safety take priority over exact trajectory tracking. This is consistent with the established stability of the underlying VDC/EMLA controller while providing an explicit safety guarantee for the prescribed workspace and actuator-stroke constraints.

\section{Validation and Results}

\subsection{Validation Setup}
The proposed control architecture is evaluated in MATLAB/Simulink using a Simscape Multibody model of the six-DoF all-electric HDRM. All controllers operate with a sampling time of \(0.001\)~s. To ensure a consistent comparison, all methods use the same plant model and unchanged inner VDC/EMLA controller; the corresponding details are provided in \cite{bahari2026all,barjini2025surrogate}.

\subsubsection{Trajectory and Safety Parameters}

\begin{figure}[t]
    \centering
    \includegraphics[width=0.5\columnwidth]{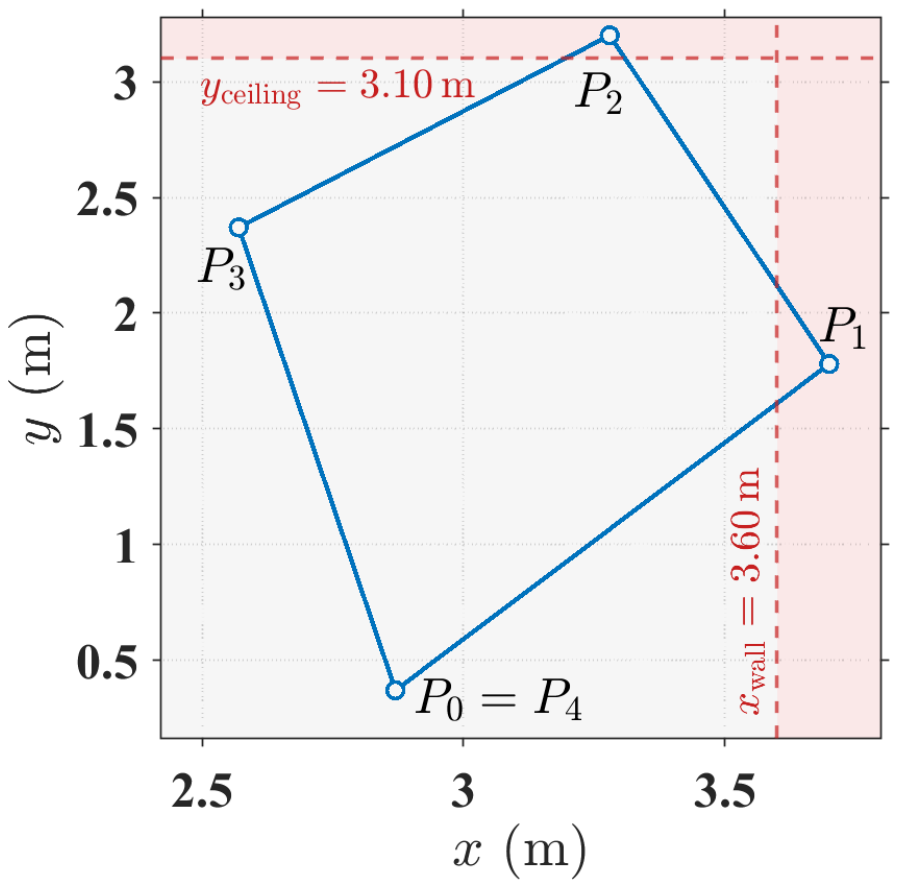}
    \caption{Commanded HDRM motion in the \(xy\)-plane together with the physical workspace boundaries.}
    \label{fig:traj_xy}
\end{figure}

Figure~\ref{fig:traj_xy} illustrates the commanded motion in the \(xy\)-plane together with the physical workspace boundaries. A 120-s Cartesian trajectory, divided into four 30-s phases, is designed to activate different safety constraints. Starting from \([2.87,0.37,0.20]^T\) m, the commanded waypoints are
\begin{equation}
\begin{aligned}
\vect P_1&=[3.70,1.78,0.20]^T,\quad
\vect P_2=[3.28,3.20,0.20]^T,\\
\vect P_3&=[2.57,2.37,0.20]^T,\quad
\vect P_4=[2.87,0.37,0.20]^T\ \mathrm{m}.
\end{aligned}
\label{eq:waypoints}
\end{equation}
The first waypoint extends beyond the safe side-wall limit, the second exceeds the safe ceiling boundary, and the third is selected to challenge the closed-chain EMLA stroke limits. The final waypoint returns the manipulator to its initial safe pose. These intentionally infeasible references are introduced to verify that the proposed safety framework prioritizes constraint satisfaction over exact tracking.

Table~\ref{tab:params} summarizes the safety and VIC parameters. The wall and ceiling are located at \(x_{\mathrm{wall}}=3.60\) m and \(y_{\mathrm{ceiling}}=3.10\) m, with \(d_{\mathrm{env}}=0.05\) m. Both EMLAs have physical stroke limits of \(0.05\)--\(0.45\) m and a safety margin \(d_{\mathrm{stroke}}=0.01\) m. All CBF gains are set to \(\gamma_i=2~\mathrm{s}^{-1}\). The robust bound \(\bar e_v\) is defined as 1.2 times the maximum measured \(\|\vect e_v\|_2\) and remains fixed during the robust simulations.

\begin{table}[t]
\caption{Main VIC, Safety, and Contact Parameters}
\label{tab:params}
\centering
\begin{tabular}{@{}p{0.52\columnwidth}p{0.38\columnwidth}@{}}
\toprule
Parameter & Value \\
\midrule

Physical wall / ceiling &
\(3.60/3.10\) m \\

Environmental margin \(d_{\mathrm{env}}\) &
\(0.05\) m \\

Safe wall / ceiling &
\(3.55/3.05\) m \\

VIC warning distance \(d_{\mathrm{warn}}\) &
\(0.40\) m \\

Physical EMLA stroke range &
\([0.05,\,0.45]\) m \\

Stroke safety margin \(d_{\mathrm{stroke}}\) &
\(0.01\) m \\

Safe EMLA stroke range &
\([0.06,\,0.44]\) m \\

CBF gains \(\gamma_i\) &
\(2~\mathrm{s}^{-1}\) \\

Robust velocity bound \(\bar e_v\) &
\(0.0192~\mathrm{rad/s}\) \\

Baseline gain \(\lambda_p\) &
\(10~\mathrm{s}^{-1}\) \\

Prediction horizon \(T_p\) &
\(0.30\) s \\

VIC stiffness \(K_{\mathrm{far}}/K_{\mathrm{near}}\) &
\(20000/3000\) N/m \\

VIC damping \(D_{\mathrm{far}}/D_{\mathrm{near}}\) &
\(2000/2500\) N\,s/m \\

Force thresholds \(F_{\mathrm{on}}/F_{\mathrm{full}}\) &
\(500/2000\) N \\

Contact stiffness \(k_e\) &
\(1\times10^{6}\) N/m \\

Contact damping \(c_e\) &
\(1\times10^{3}\) N\,s/m \\

Transition width \(w\) &
\(1\times10^{-4}\) m \\

\bottomrule
\end{tabular}
\end{table}

\subsubsection{Control Configurations and Contact Scenario}
The four configurations in Table~\ref{tab:methods} isolate the contribution of each control layer. Method 0 uses the established VDC-based all-electric HDRM controller \cite{bahari2026all,barjini2025surrogate}. Method~1 activates VIC but bypasses the CBF filter. Method~2 uses the CBF filter with the original fixed task-space error feedback, and Method~3 combines VIC and robust CBF filtering. 

\begin{table}[t]
\caption{Compared Control Configurations}
\label{tab:methods}
\centering
\begin{tabular}{@{}clcc@{}}
\toprule
Method & Description & VIC & CBF \\
\midrule
0 & Original VDC & -- & -- \\
1 & VIC + VDC & \checkmark & -- \\
2 & CBF + VDC & -- & \checkmark \\
3 & VIC + CBF + VDC & \checkmark & \checkmark \\
\bottomrule
\end{tabular}
\end{table}

For the physical-contact tests, the existing wall and ceiling are treated as the environmental contact surfaces and modeled using the Simscape Multibody spatial-contact formulation. The normal contact interaction is represented by a smooth spring--damper penalty law,
\begin{equation}
F_n=s(\delta,w)\left(k_e\delta+c_e\dot{\delta}\right),
\label{eq:contact}
\end{equation}
where \(\delta\) denotes the penetration depth, \(k_e\) and \(c_e\) are the normal contact stiffness and damping coefficients, respectively, and \(s(\delta,w)\) is a smooth transition function with width \(w\) that regularizes contact activation. Tangential friction is neglected. The resulting wall and ceiling contact-force components are combined to form the Cartesian environment-interaction force \(\vect F_{\mathrm{env}}\), which is supplied to the VIC layer.

\begin{table*}[t]
\caption{Quantitative comparison of the four control configurations.}
\label{tab:results}
\centering
\scriptsize
\begin{tabular}{@{}lccccccc@{}}
\toprule
Method & $e_{\mathrm{RMSE}}$ & $\min h_{\mathrm{env}}$ & $\min h_{\mathrm{EMLA}}$ & $\max\|\vect e_v\|_2$ & $J_{\mathrm{CBF}}$ & Peak/RMS $\|\vect F_{\mathrm{env}}\|_2$ & Safe? \\
 & [m] & [m] & [m] & [rad/s] & [rad$^2$/s] & [N] & \\
\midrule
VDC & 0.0417 & -0.0510 & -0.0327 & -- & -- & 20773.5/12204.3 & No \\
VIC+VDC & 0.0202 & -0.0330 & -0.0325 & -- & -- & 1539.4/448.4 & No \\
CBF+VDC & 0.0043 & 0.0121 & 0.0014 & 0.0154 & 89.37 & 0.0/0.0 & Yes \\
VIC+CBF+VDC & 0.0070 & 0.0121 & 0.0014 & 0.0154 & 80.16 & 0.0/0.0 & Yes \\
\bottomrule
\end{tabular}
\vspace{0.5mm}\par\centering\scriptsize
The Cartesian RMSE is evaluated over the common feasible set \(\mathcal T_f\); negative barrier values indicate constraint violation.
\par
\end{table*}

Controller performance is evaluated using Cartesian RMSE over the common feasible set \(\mathcal T_f\), minimum CBF values and constraint violations, the joint-velocity realization error \(\|\vect e_v\|_2\), CBF intervention, and contact-force magnitude. The feasible-set RMSE avoids penalizing safety-induced rejection of intentionally infeasible references. The CBF intervention is quantified by
\begin{equation}
J_{\mathrm{CBF}}
=
\int_0^T
\norm{
\dot{\vect{\Theta}}_{\mathrm{safe}}
-
\dot{\vect{\Theta}}_{\mathrm{nom}}
}_2^2\,\dd t ,
\label{eq:jcbf}
\end{equation}
where a smaller value indicates less modification of the nominal joint-velocity command. Robustness is assessed by comparing the measured \(\|\vect e_v\|_2\) with the prescribed bound \(\bar e_v\). For the contact test, peak and RMS values of \(\|\vect F_{\mathrm{env}}\|_2\) are additionally reported.

\subsection{Results and Discussion}

The first two trajectory phases deliberately command positions beyond the prescribed wall and ceiling safety limits, while the third challenges the closed-chain EMLA stroke range. Methods~0 and~1 do not impose hard barrier constraints and therefore violate the admissible workspace and actuator limits. In contrast, Methods~2 and~3 project the nominal joint-velocity command onto the common CBF-admissible set, thereby prioritizing safety over exact tracking of infeasible references.

\begin{figure}[t]
\centering
\includegraphics[width=0.9\columnwidth]{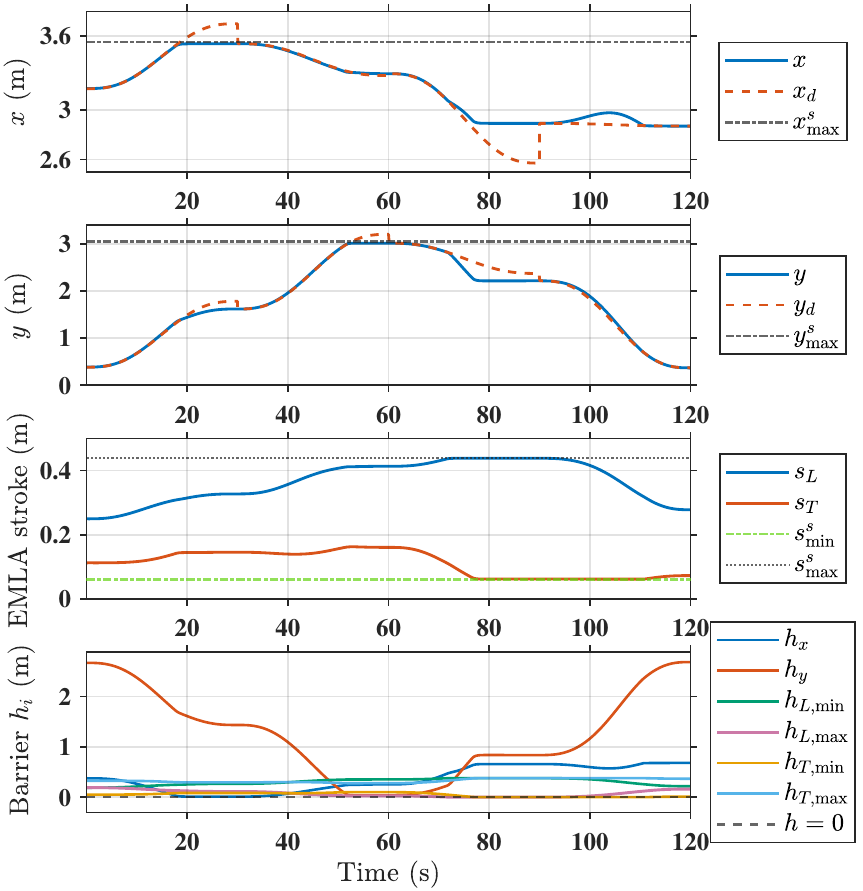}
\caption{Validation of the proposed VIC--CBF--VDC architecture~(Method~3). From top to bottom: end-effector motion in the \(x\)-direction, end-effector motion in the \(y\)-direction, lift and tilt EMLA strokes, and the six workspace and EMLA CBFs.}
\label{fig:constraint_results}
\end{figure}

Figure~\ref{fig:constraint_results} illustrates the behavior of the proposed Method~3. Although the desired Cartesian trajectory exceeds the prescribed workspace boundaries, the actual motion remains within \(x_{\max}^{s}=3.55\)~m and \(y_{\max}^{s}=3.05\)~m. Similarly, the lift and tilt EMLA strokes remain within the safe interval \(0.06\)--\(0.44\)~m. All six CBFs remain nonnegative, with minimum workspace and EMLA barrier values of \(0.0121\)~m and \(0.0014\)~m, respectively.

Table~\ref{tab:results} provides the quantitative comparison of all four methods. The Cartesian RMSE is evaluated over the common feasible set \(\mathcal T_f\), such that intentional rejection of unsafe references is not interpreted as tracking failure. Without CBF enforcement, Methods~0 and~1 exhibit maximum constraint violations of \(0.0510\)~m and \(0.0330\)~m, respectively. VIC nevertheless substantially improves the physical-interaction response: relative to the VDC baseline, Method~1 reduces the peak environment-interaction force from \(20.77\)~kN to \(1.54\)~kN and the RMS force from \(12.20\)~kN to \(0.448\)~kN, corresponding to reductions of \(92.59\%\) and \(96.33\%\), respectively. These results demonstrate that VIC can strongly mitigate contact severity, but does not by itself guarantee safety.

Both CBF-based methods preserve the prescribed safe sets and prevent physical contact with the wall and ceiling, resulting in zero measured environment-interaction force. The maximum measured velocity-realization error is \(0.0154~\mathrm{rad/s}\) for both Methods~2 and~3 and remains below the prescribed robust bound \(\bar e_v=0.0192~\mathrm{rad/s}\), supporting the robustness assumption used in the CBF tightening. Furthermore, introducing VIC reduces the CBF intervention metric from \(J_{\mathrm{CBF}}=89.37\) for Method~2 to \(80.16~\mathrm{rad^2/s}\) for Method~3, corresponding to a \(10.30\%\) reduction. Thus, VIC shapes the nominal motion toward a command that is more compatible with the safety constraints, while the robust CBF layer retains responsibility for hard constraint enforcement.

The results therefore separate the roles of the two added control layers. VIC significantly reduces the severity of physical interaction when contact is allowed, but does not provide a formal safety guarantee. The robust CBF filter prevents workspace and actuator-limit violations, while their combination in Method~3 preserves the same safety margins and requires less corrective intervention from the CBF filter.

\section{Conclusion}

This paper presented a VIC--CBF--VDC architecture for safe physical interaction of all-electric EMLA-driven HDRMs. Risk-aware VIC shapes the nominal task-space behavior according to predicted proximity and measured interaction force, while a robust joint-velocity CBF filter simultaneously enforces Cartesian workspace and closed-chain EMLA stroke constraints. The CBF formulation explicitly accounts for the finite velocity-realization accuracy of the underlying VDC/EMLA controller.
Simulation results demonstrated the complementary roles of VIC and CBF. VIC alone reduced the peak and RMS environment-interaction forces by \(92.59\%\) and \(96.33\%\), respectively, relative to the VDC baseline, but did not prevent constraint violations. In contrast, both CBF-based configurations maintained nonnegative barrier functions, with minimum workspace and EMLA margins of approximately \(0.0121\)~m and \(0.0014\)~m, respectively, while the maximum measured velocity-realization error of \(0.0154~\mathrm{rad/s}\) remained below the prescribed robust bound. Combining VIC with the robust CBF retained these safety guarantees while reducing the CBF intervention metric by \(10.30\%\) relative to CBF+VDC. These results show that compliance and formal safety can be treated as complementary functions within the same hierarchical architecture. 
Future work will extend the constraint set to moving obstacles and neighboring manipulators and validate the proposed framework on the full physical platform.

\section*{Acknowledgment}
Portions of the manuscript text were prepared and revised with the assistance of OpenAI ChatGPT (GPT-5.6 Sol, 2026). All technical content and results were reviewed and verified by the authors.

\bibliographystyle{IEEEtran}
\bibliography{references}

\end{document}